\documentclass[lettersize,journal]{IEEEtran}
\usepackage{amsmath,amsfonts}
\usepackage{algorithmic}
\usepackage{algorithm}
\usepackage{array}
\usepackage[caption=false,font=normalsize,labelfont=sf,textfont=sf]{subfig}
\usepackage{textcomp}
\usepackage{stfloats}
\usepackage{verbatim}
\usepackage{graphicx}
\usepackage{cite}
\usepackage{mfirstuc}

\usepackage{hyperref}
\hypersetup{colorlinks=true,
	linkcolor=black, 
	urlcolor=teal, 
    citecolor=black} 

\usepackage[x11names]{xcolor}
\usepackage{pgfplotstable}
\usepackage{glossaries}

\newacronym{3gpp}{3GPP}{3$^\text{rd}$ generation partnership project}
\newacronym{5g}{5G}{5$^\text{th}$ generation}
\newacronym{ai}{AI}{artificial intelligence}
\newacronym{aoa}{AoA}{angle of arrival}
\newacronym{ap}{AP}{access point}
\newacronym{cnn}{CNN}{convolutional neural network}
\newacronym{csi}{CSI}{channel state information}
\newacronym{dot}{dot}{radio dot system}
\newacronym{ecdf}{ECDF}{empirical cumulative distribution function}
\newacronym{flops}{FLOPs}{FLoating OPerations}
\newacronym{iq}{I/Q}{in-phase and quadrature}
\newacronym{isac}{ISAC}{integrated sensing and communication}
\newacronym{ml}{ML}{machine learning}
\newacronym{ofdm}{OFDM}{orthogonal frequency division multiplexing}
\newacronym{slam}{SLAM}{simultaneous localization and mapping}
\newacronym{ssb}{SSB}{synchronization signal block}
\newacronym{ue}{UE}{user equipment}
\newacronym{usrp}{USRP}{universal software radio peripheral}

\usepackage{csvsimple}
\usepackage{tabularx, booktabs}

\usepackage{tikz,bm}
\usepackage{pgfplots}
\usetikzlibrary{calc}
\usepgfplotslibrary{fillbetween}
\pgfplotsset{compat=1.16}

\usepackage{booktabs}

\newcommand\remembertext[2]{
  \immediate\write\@auxout{\unexpanded{\global\long\@namedef{mytext@#1}{#2}}}%
  {\color{blue} #2}%
}

\newcommand\recalltext[1]{%
  \ifcsname mytext@#1\endcsname
    \@nameuse{mytext@#1}%
  \else
    ``???''
  \fi
}

\begin{document}
\title{FAWN: A Multi-Encoder Fusion-Attention Wave Network for Integrated Sensing and Communication Indoor Scene Inference}

\author{Carlos Barroso-Fernández, Alejandro Calvillo-Fernandez, Antonio de la Oliva, and Carlos J. Bernardos.
\thanks{This work has been submitted to the IEEE for possible publication. Copyright may be transferred without notice, after which this version may no longer be accessible.
}}




\maketitle

\begin{abstract}
Integrated sensing and communication (ISAC) emerges as a complementary solution to traditional radar sensing by reusing communication signals to sense the environment, extending perception beyond the line-of-sight. However, traditional methods ignore the presence of different radio technologies that can capture supplementary information. 
In this work, we introduce FAWN, a Fusion-Attention Wave Network designed to infer indoor scenes through \emph{passive} ISAC, merging information from multiple technologies. FAWN employs a multi-encoder architecture that fuses heterogeneous channel state information (CSI) from different devices, enabling the detection and localization of entities without requiring them to be network users. Experiments in a warehouse testbed equipped with commercial Wi-Fi 6 and 5G NR deployments show that FAWN achieves 96\% classification accuracy and sub-meter localization precision ($\le$0.6 m in 84\% of cases), outperforming state-of-the-art models without the need to synchronize emitted signals. Results confirm the potential of multi-technology fusion for \emph{passive} ISAC to enhance perception robustness and spectral efficiency in next-generation robotic and cyber-physical systems.

\end{abstract}










\section{Introduction}
\label{sec:intro}

\IEEEPARstart{T}{he} new generation of mobile communications, i.e., 6G, is founded with a new perspective than its predecessors. Previous generations mainly devote their improvements to achieve enhanced mobile broadband; however, 6G also brings a set of new services beyond communications. One of those is wide area multi-dimensional sensing~\cite{3rd_generation_partnership_project_3gpp_study_2026,meng_cooperative_2025}, which offers spatial information about the physical environment and objects (unconnected and connected to the network).

\capitalisewords{\gls{isac}} proposes to reuse the same radio link designed for communication functions to sense the environment, leveraging their wide coverage. Some existing approaches employ \textit{active sensing} over wireless communication, where dedicated signals are used to map the surroundings~\cite{wen_survey_2024}. While effective, active sensing increases spectral congestion by introducing new dedicated transmissions into an already crowded radio environment. In contrast, \textit{passive sensing} emerges as a cost-effective solution that reuses the existing communication signals passively to avoid interfering with current transmissions~\cite{wen_survey_2024}.


Passive sensing is accomplished by analyzing the \gls{csi}, which is calculated comparing the received signal with a reference. 
As the \gls{csi} captures the degradation of the signal, it also contains information of how obstacles have modified it, e.g., a person, who can be detected even if they are not carrying any device connected to the network. But this technology also has its drawbacks. The fact that it is not dedicated to sensing tasks may derive in a lack of precision, and poor performance when the problem grows in complexity. 

In this work, we explore the use of passive sensing to infer the environmental context in a human-robot collaborative space by leveraging the existing wireless infrastructure. 
Authors in the literature use invariant reference signals such as Wi-Fi beacons~\cite{he_robust_2023,hussain_wisigpro_2024} or \acrfull{5g} \glspl{ssb}~\cite{dwivedi_5g-based_2024,wypich_5g-based_2025} to get the \gls{csi}. 
However, existing works analyze signals from one or the other technology~\cite{he_robust_2023,hussain_wisigpro_2024,dwivedi_5g-based_2024,wypich_5g-based_2025}, overlooking the complementary information that brings different nodes' perspective. 
Instead of relying on a single observation point, we integrate measurements from multiple network nodes (possibly, from different technologies), namely Wi-Fi and \gls{5g}, to obtain complementary perspectives of the same scene. This multi-node, multi-technology view increases robustness against spatial blind spots and technology-specific failure regions, improving inference accuracy compared to single-node baselines. Specifically, our contributions are the followings:

\begin{itemize}
    \item We present {FAWN}, a multi-encoder Fusion-Attention Wave Network that uses heterogeneous technologies for indoor presence detection and localization, which performs satisfactorily under critical cases without the information of dedicated sensors.
    \item We effectively merge the feature information from multiple sources without the need of preserving channel coherence, which requires highly accurate and challenging time synchronization.
    \item We validate the framework by building a realistic testbed, deploying two passive receivers to listen the signals from a commercial Wi-Fi \acrfull{ap} and a commercial \gls{5g} indoor \gls{dot} from Ericsson (see Fig.~\ref{fig:model}); the objective is to infer the presence and position of a robot and a person. Notice that the sensed element does not need to implement any electronic system, e.g., people without a mobile phone.
\end{itemize}

The rest of the article is organized as follows. Sec.~\ref{sec:background} details some background concepts and literature articles. 
Sec.~\ref{sec:testbed} presents our testbed and evaluate related works to derive our motivation.
Sec.~\ref{sec:architecture} describes the characteristics of our proposed model, and 
Sec.~\ref{sec:results} evaluates the results of our proposed model over our testbed, in comparison with ablation works and state-of-the-art benchmarks.
Finally, Sec.~\ref{sec:futureworks} summarizes and concludes this article.

\section{Indoor passive ISAC for robotics}
\label{sec:background}

While on-board sensors remain the primary source of environmental awareness for mobile robots, their coverage is limited to its line of sight. This fact constrains the robot reaction time and restricts its action plan, inducing sub-optimal performance and, most importantly, compromising the safety of people inside collaborative scenarios.
In this sense, \gls{isac} emerges as a promising candidate to enhance environmental awareness. 

\gls{isac} envisions a unified framework in which the same hardware and signals used for communication are also exploited for sensing tasks. By combining these functionalities, \gls{isac} delivers benefits such as improved spectral efficiency, reduced hardware costs, and better situational awareness in non-idealistic environments. 
This not only enhances the practicality and scalability of sensing deployments, but also enables pervasive sensing in environments that are already covered by modern wireless technologies. 

Passive sensing refers to the technique of extracting meaningful information from the environment, like detecting the presence, motion, or location of objects, by analyzing signals that were not specifically transmitted for sensing purposes. 
Unlike active sensing systems that emit dedicated probing signals, passive sensing leverages existing ambient signals without requiring any additional transmission. 
Compared against its active counterpart, passive sensing reduces energy consumption, minimizes interference, and preserves spectral efficiency, however, it also poses significant challenges as lack of performance, or emitters synchronization.


\subsection{Sensing with indoor wireless communication signals}

The IEEE 802.11 standard~\cite{80211_wg_-_wireless_lan_working_group_standard_2024} is available in almost all buildings. Originally designed for data connectivity, Wi-Fi operates primarily in the 2.4$\,$GHz and 5$\,$GHz unlicensed bands, and more recently, also in the 6$\,$GHz band since Wi-Fi 6E. The ability of Wi-Fi signals to penetrate walls, reflect off objects, and exhibit multi-path propagation has made it particularly suitable for sensing applications.
Wi-Fi-based sensing techniques typically exploit \gls{csi}, which provides detailed information about the \gls{iq} components of the subcarriers in the \gls{ofdm}-based transmissions and enables precise tracking of environmental changes. These characteristics have been used in applications ranging from human activity recognition and fall detection~\cite{he_robust_2023,hussain_wisigpro_2024} to localization and occupancy estimation~\cite{arun_p2slam_2022,li_wifi-csi_2024}.
The Wi-Fi standardization community has recognized the potential of these applications, establishing a specific task group to enable them within the IEEE 802.11bf amendment.


\gls{5g} brings new opportunities for sensing in both indoor and outdoor environments. The \gls{3gpp}, responsible for \gls{5g} standardization, has introduced major enhancements in terms of latency, reliability, and bandwidth, all of which are essential for high-resolution sensing. 
Specifically, small cells provide high signal quality and spatial resolution due to their limited coverage area, making them suitable for detailed sensing tasks in indoor spaces such as smart buildings, factories, and retail environments. Note that outdoor antennas, same as connectivity, may have problems to perform indoor sensing.
Moreover, given to its importance, \gls{3gpp} has decided to explicitly integrate and support \gls{isac} as a native core feature in the upcoming generations, starting from Beyond 5G in Release 18~\cite{3rd_generation_partnership_project_3gpp_study_2026}.
These initiatives aim to standardize waveform designs, signal processing techniques, and network architectures that support simultaneous communication and sensing. 

\subsection{Related works}
\label{subsec:relatedWorks}


As aforementioned, Wi-Fi is the most commonly deployed network in indoor environments, and as a result, it has been extensively studied in the literature.
Authors of \cite{li_wifi-csi_2024} propose \textit{FewSense}, a Wi-Fi passive‐tracking approach that estimates Doppler speed by taking time-domain differences of consecutive \gls{csi} samples. Many other approaches opt to use \gls{ml}, specifically, \glspl{cnn} have shown great performance due to their usage of the convolution to process the signal. The framework presented in~\cite{he_robust_2023} proposes a Wi-Fi passive sensing method based on CSI, named CNN-ABLSTM, 
with an attention mechanism to recognize human activities. Another example is \textit{WiSigPro}~\cite{hussain_wisigpro_2024}, an encoder-only Transformer~\cite{vaswani_attention_2017} for CSI-based Wi-Fi passive human activity recognition.

It is worth mentioning the \gls{slam} use case, in which a signal dedicated to localize a connected station is reused to map the environment. In this line, \textit{P2SLAM}\cite{arun_p2slam_2022} is a Wi-Fi-based indoor \gls{slam} system that interprets \glspl{ap} as unknown landmarks and fuses two-way bearing estimates, derived from CSI as \glspl{aoa}, with wheel-odometry to jointly localize the robot and map the \glspl{ap}. Authors of~\cite{picazo_waveslam_2023} present a \gls{slam} solution, this time leveraging the millimeter wave band, to improve the existing mechanisms within the robot, such us LiDARs. 




    


Recently, with the advent of \gls{5g} indoor deployments, this technology has also started to be applied in similar scenarios \cite{arun_p2slam_2022}, but is still a growing field. Most of the works rely on signal processing properties to infer what is in the environment, but its accuracy and usage is limited~\cite{nataraja_integrated_2025}. Recent solutions like in~\cite{dwivedi_5g-based_2024} rely on \gls{ml} to improve the performance, reaching high accuracy with low-cost solutions under simple scenarios.  Complementarily,~\cite{wypich_5g-based_2025} uses \gls{5g} CSI leveraging a Fast-Fourier Transform (FFT) to enable low-complexity passive radar with meter-level range resolution.

\smallskip

As we have seen, the environmental perception using \gls{isac} is a far from a trivial task. In fact, current models from the literature rely on restrictions over the scenario (e.g., sender and receiver must have line of sight) to be able to find closer expressions for the degradation of the signal over a wireless channel. Furthermore, the solutions from the literature use one of the two technologies (Wi-Fi or \gls{5g}) to passively sense the environment. As each passive sensing modality 
typically operates under different protocols, sampling rates, noise characteristics, and spatial or temporal resolutions, we combine information from multiple sources to enhance the environmental perception performance.




\section{Mixing different sensors}
\label{sec:testbed}

\begin{table*}[t]
  \centering
  \caption{Comparison of Wi-Fi and \gls{5g}-based passive sensing methods. Last three columns (Position accuracy, Scenario accuracy and Limitations of the framework) refer to its implementation within our testbed.}
  \label{tab:sota}
  \setlength{\tabcolsep}{3pt}
  \renewcommand{\arraystretch}{1.15}
  \scriptsize
  \begin{tabularx}{\textwidth}{@{}p{.7cm}p{1.1cm}p{0.5cm}p{1.9cm}p{1.9cm}p{3.9cm}p{2cm}p{1.1cm}p{3.1cm}@{}} 
    \toprule
    \textbf{Work} & \textbf{Wi-Fi} & \textbf{\gls{5g}} & \textbf{Use Case} & 
    \begin{tabular}{l}
        \textbf{Claimed} \\
        \textbf{results}
    \end{tabular}
    & \textbf{Methodology} &
    \begin{tabular}{l}
        \textbf{Position} \\
        \textbf{accuracy}
    \end{tabular} & \begin{tabular}{l}
        \textbf{Scenario} \\
        \textbf{accuracy}
    \end{tabular} & 
        \textbf{Limitations of the framework}  \\
    \midrule
    

    \cite{li_wifi-csi_2024} 
    & Doppler 
    & – &
    Tracking & $<$ 0.34$\,$m in 50\textsuperscript{th} \newline percentile  &
    \textit{FewSense} subtract two CSI samples \newline to obtain the Doppler speed.
    &
    11\% &
    56.9\%  & 
    Only moving targets. 
    \\ \addlinespace[3pt]
    
    \cite{he_robust_2023} 
    & CSI 
    & – &
    Human activity \newline recognition & 97\% accuracy &
    CNN-ABLSTM with attention. &
    69\%  &
    86.7\%  & Person must block the line of sight of the stations. \\ \addlinespace[3pt]

    \cite{hussain_wisigpro_2024} 
    & CSI 
    & – &
    Human activity \newline recognition & 98\% accuracy &
    Multi-head attention mechanisms and \newline positional encoding. &
    4\% &
    43.5\%  & Person must block the line of sight of the stations. \\ \addlinespace[3pt]
    
    \cite{dwivedi_5g-based_2024} 
    & – & CSI 
    & Human activity \newline recognition & 98.5\% accuracy&
    Modified  LeNet-5~\cite{lecun_gradient-based_2002} network. &
    38\% &
    87.9\% & Person must block the line of sight of the stations. \\ \addlinespace[3pt]
    
    \cite{wypich_5g-based_2025}  
    & – & CSI &
    Target detection & $\approx$ 1.5$\,$m in 50\textsuperscript{th} \newline percentile &
    FFT to create a speed-range map.
    &
    2\% &
    13\%  & Reinforced with dedicated \newline sensing signals. 
    \\ \addlinespace[3pt]

    \textit{\gls{5g} only}
    & – & CSI &
    Presence detection \newline and localization & $\approx$ 0.6$\,$m in 54\textsuperscript{th} \newline percentile &
    \gls{5g} encoder attention based. & 34\% &
    67.3\%  & Lack of coverage.
    \\ \addlinespace[3pt]

    \textit{Wi-Fi only}
    & CSI & – &
    Presence detection \newline and localization & $\approx$ 0.6$\,$m in 69\textsuperscript{th} \newline percentile &
    Wi-Fi encoder attention based. & 74\% &
    85.5\%  & Lack of coverage.
    \\ \addlinespace[3pt]

    \textit{FAWN}
    & CSI & CSI &
    Presence detection \newline and localization & $\approx$ 0.6$\,$m in 84\textsuperscript{th} \newline percentile &
    Multi-encoder, attention based \newline aggregation, and single-decoder. & 86\% &
    96\%  & Generalization over unseen scenarios
    \\

    \bottomrule
  \end{tabularx}
\end{table*}

\begin{figure}
\centering
\begin{tikzpicture}

\definecolor{pubuLight}{RGB}{236, 231, 242}  
\definecolor{pubuMed}{RGB}{166, 189, 219}    
\definecolor{pubuDark}{RGB}{43, 140, 190}    

\begin{axis}[
    ybar=3pt, 
    bar width=15pt, 
    width=\columnwidth, 
    height=0.5\columnwidth, 
    enlarge x limits=0.3, 
    legend style={
        at={(1,1)}, 
        anchor=north east, 
        legend columns=1, 
        font=\footnotesize,
        draw=none,
    },
    ylabel={Accuracy},
    ymin=0.28, ymax=.92, 
    symbolic x coords={Linear Fusion, Wi-Fi only, 5G only},
    xtick=data,
    x tick label style={
        font=\footnotesize,
        text width=3cm,
        align=center
    },
    nodes near coords,
    every node near coord/.append style={
        font=\tiny,
        rotate=60, 
        anchor=south west,
        /pgf/number format/fixed,
        /pgf/number format/fixed zerofill,
        /pgf/number format/precision=3
    },
    axis lines*=left,
    ymajorgrids=true,
    grid style={dashed, gray!30}
]

\addplot[fill=pubuLight, draw=darkgray] coordinates {
    (Wi-Fi only, 0.748) 
    (5G only, 0.300) 
    (Linear Fusion, 0.816)
};

\addplot[fill=pubuMed, draw=darkgray] coordinates {
    (Wi-Fi only, 0.768) 
    (5G only, 0.410) 
    (Linear Fusion, 0.914)
};

\addplot[fill=pubuDark, draw=darkgray] coordinates {
    (Wi-Fi only, 0.756) 
    (5G only, 0.346) 
    (Linear Fusion, 0.857)
};

\legend{Robot pos, Person pos, Both pos}

\end{axis}
\end{tikzpicture}%
\caption{Accuracy comparison per sensor configuration.}%
\label{fig:accuracy_bar_chart}%
\end{figure}

Because of the strict assumptions when modeling scenarios, accuracy from literature's solutions drops when applied to more general and complex environments. We realized this fact when trying to solve a realistic sensing problem: identifying the position of human operators and mobile robots within a room where both coexist.

The room is a $5.4\times 6\,$m rectangle, with 3$\,$m of height. We leverage signals from two already deployed communication networks nodes: one commercial \gls{ap} Turris Omnia Wi-Fi~6 equipped with AW7915-NP1 network card, and one commercial \gls{5g} \gls{dot} from Ericsson. They are located at 2 and 3 meters from the ground, and emit over $40\,$MHz width channels centered at $5.5\,$ and $3.62\,$GHz, respectively. To listen the signals, we deploy passive receptors for each technology using USRPs B210 software-defined radio, which captures them and extract \gls{csi} from Wi-Fi beacons and \gls{5g} \glspl{ssb} (note that it can be used other reference signals). The devices uses broadband monopole antennas covering their respective bands, operating in single-input single-output mode.
We collect labeled samples of mixed human and robot activity, containing \gls{csi} for both wireless networks and the position of the robot and/or person. 
The steps of the moving entities coincides with a $9\times 10$ grid of 0.6$\,$m squares. In total, we collected more than $17\,000$ samples. The ratio of the training, validation and test sets are, respectively, 80/10/10\%, and, to maintain statistical significance, experiments are repeated with different instances.

Using this data, we compare multiple works from the literature reviewed in Sec.~\ref{subsec:relatedWorks}. To do so, we have adapted and deployed them in our scenario. The results are summarized in Table~\ref{tab:sota}, which presents the technology, use case, claimed results, methodology, results over our testbed (position and scenario accuracy), and limitations of each article. Overall, we can see the drop of precision of their claimed results against the performance over our testbed. As mentioned, this effect is mainly driven by the assumptions over the environment, which translates into lack of generalization~\cite{zhang_lessons_2026}.


The more challenging a task is, the more data \gls{ml} models will need. Our idea is to reuse multiple sources of communication signals to augment the feature dimension, increasing the amount of information the model receives.
To prove its validity, we develop three complementary linear \gls{ml} models. \textit{Wi-Fi Only}, uses only the information from Wi-Fi \gls{csi}; \textit{5G Only}, uses only the information from \gls{5g} \gls{csi}; \textit{Linear Fusion} uses both, and merge the information with another linear layer.


Fig.~\ref{fig:accuracy_bar_chart} shows the accuracy comparison of these three models across the different localization tasks (Robot, Person, and a combination of both). As observed, relying solely on \gls{5g} data yields the lowest performance, whereas the Wi-Fi data provides a much stronger independent baseline. Most importantly, the \textit{Linear Fusion} model demonstrates that by merging both \gls{csi} sources, the model overcomes the individual limitations of each node and achieves the highest accuracy across all three positioning tasks, peaking at over 91\% accuracy for the person's position.

\smallskip
These results demonstrate that merging information from different sources enriches environmental knowledge, unlocking the possibility of sensing under more complex scenarios.

\section{Improving Feature Fusion: An attention approach}

\label{sec:architecture}

\begin{figure*}
    \centering
    \includegraphics[width=\linewidth]{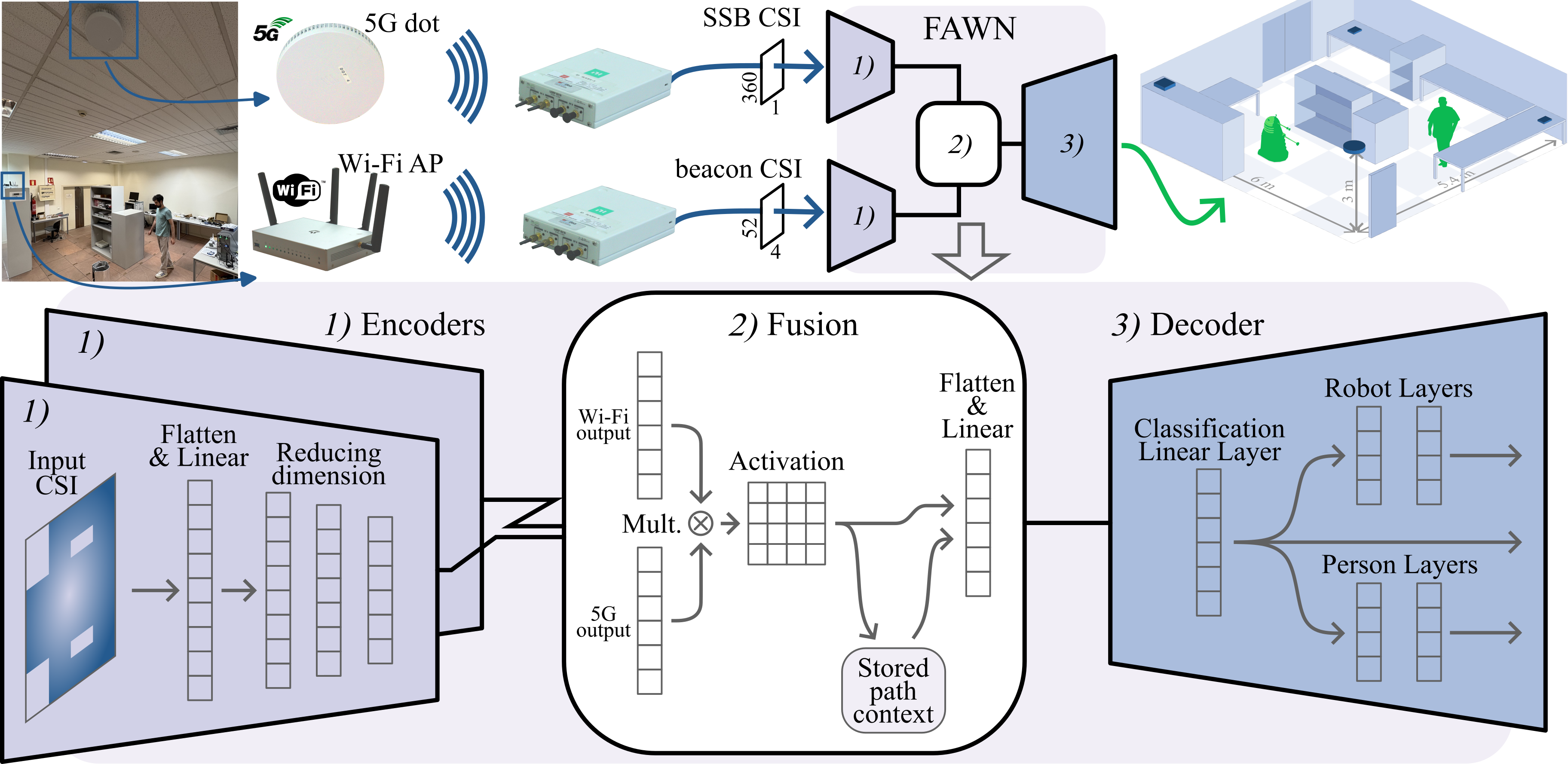}
    \caption{The flow of how \gls{csi} is extracted from the real environment (top left picture) using the USRPs acting as passive receivers of \gls{5g} \glspl{ssb} and Wi-Fi beacons. Then, FAWN (light purple background) use them to infer where the person and robot are (top right). Bottom part depicts the layers of the three building blocks of FAWN: \textit{1)} the encoders, \textit{2)} the sensor fusion, and \textit{3)} the decoder.}
    \label{fig:model}
\end{figure*}



Driven by the experimental validation presented in the previous section, we aim to maximize the information extracted from each technology, weighting the features in function of their impact and amount of latent information. Attention over \gls{ml} refers to a set of mechanisms which stand out by precisely doing that kind of weighting~\cite{vaswani_attention_2017}. 


\textit{FAWN}, a multi-encoder Fusion-Attention Wave Network, can be understood as 
an aggregating system, where the information from multiple encoders is fused and processed by a single decoder. 
In the classical example, each encoder handles a different language, and the decoder turns their shared representation into a translation; here, each encoder process the \gls{csi} from a single device. For our testbed, as we have two different radio fingerprints, one encoder processes the information from the \gls{5g} emitter, while the other one handles Wi-Fi. The decoder gets the merged embedded information from the two technologies and uses it to estimate what, and where, is inside the room. This is depicted in Fig.~\ref{fig:model}--(top), where we can see how \textit{FAWN} processes the data generated by the \gls{5g} \gls{dot} and the Wi-Fi \gls{ap} from the testbed, each with a different encoder, and outputs the type and location of the elements inside the room.

\subsection{Modular building blocks}
\label{ss:arquitectura}

Motivated by the goal of creating a modular, distributed, and scalable architecture, \textit{FAWN} assigns a separate encoder to each source of information. Specifically in our testbed, we have available a 5G \gls{dot} and a Wi-Fi \gls{ap}; for more details on our testbed, we refer the reader to Sec.~\ref{sec:testbed}. However, new sensors can be added up by clipping another encoder block without rewriting the rest of the pipeline. In the case that the scenario has a third radio fingerprint, \textit{FAWN} can add up that information to increase the environmental context.

\subsubsection{Multi-encoders}
\label{sss:encoder}

They process the \gls{csi} information to create the corresponding embeddings, which later will be jointly fused. 
Each encoder is specialized in one technology to spotlight its most important features. Additionally, they are in charge of reducing the size of the features to decrease the complexity of the problem.

The structure of each encoder begins with a flattening process and a Linear layer. Then, three additional hidden Linear layers reduce the dimension of the features to a common embedded size.
Each encoder share the same structure but differs in the input dimensions. In other words, the operations are the same, but the size of the first layers is adapted to each technology. All these layers are represented in Fig.~\ref{fig:model}--(bottom left).

\subsubsection{Sensor Feature fusion}
\label{sss:fusion}

The core design principle of \textit{FAWN} is the fusion of heterogeneous sensing-related data streams to achieve a more robust understanding of the surrounding environment. Within a scene, we can find multiple radio communication nodes using the same or different technologies. The different perspective provided by each emitter captures complementary information that might be missed by a single modality.

However, because of the basis of passive sensing, we cannot enforce all emitters to send the signals at the same time. The framework must be able to understand the time gap between the received data and weight them in function of its importance. Additionally, if the sensed object is out of the coverage area of one of the emitters, it will not report valuable information. The model must also contemplate it when merging the data from the multiple encoders, to expand its sensed range.

These abilities are acquired through the self-attention mechanism of the data fusion block, which utilizes a learnable projection matrix to retain only the most relevant features from each technology. Additionally, it is enriched with path context by concatenating previously stored states. Fig.~\ref{fig:model}--(bottom middle) shows a scheme of this process. To ensure mathematical symmetry for its operations, it requires input embeddings of an identical length from all branches. 

\subsubsection{The Decoder}
\label{sss:decoder}

The final block is a lightweight decoder, as the heavy lifting has been carried out by the previous block. It projects the information retained in the embedding vector to a feasible answer for the specific task.
The decoder is composed of five linear layers: one dedicated to what is in the room, and the other four to discretize the $x$-$y$ position of the detected presences -- see Fig.~\ref{fig:model}--(bottom right).

\smallskip

This modular approach allows to deploy FAWN in a disaggregated manner, distributing the building blocks along the  system. Each encoder may be instantiated within the source of information, not exposing the raw data. The embedded data fusion occurs as soon as possible, reducing the impact on the network. Last, the decoder is localized at the core, where the service context resides.

\subsection{Workflow}
\label{ss:workflow}

In our testbed, we have available two distinct wireless technologies. Hence, we deploy \textit{FAWN} with two parallel encoders, each with an input layer tailored to the specific shape of its corresponding signal:

\begin{itemize}
    \item \textbf{\gls{5g} \gls{ssb}}: The input shape of the encoder is a $2 \times 360 \times 4$ tensor. Respectively, 2 channels for I/Q data, 360 channels for each of the subcarriers, and 4 for each of the symbols sent.
    
    \item \textbf{Wi-Fi Beacon}: The input shape of the encoder is a $2 \times 52 \times 1$ tensor, respectively, 2 channels for I/Q data, 52 channels for each of the subcarriers, and 1 for each of the symbols sent.
\end{itemize}

Each signal begins its journey in its own dedicated encoder. They begins by flattening the tensors into vectors, followed by a linear layer. This first layer serves as the input interface, and starts capturing relationships between features. After it, three additional hidden linear layers are dedicated to reduce the size of the embedding vector to a common dimension, simplifying the complexity by pruning unnecessary information from the features. In particular, 5G encoder hidden linear layers have 384, 256, and 96 neurons; and Wi-Fi encoder ones have 256, 192, and 96. Both outputs have the same embedding dimension, which are the input for the fusion block.

In the next block, \textit{FAWN} stacks the information coming from the encoders through a self-attention mechanism. The block begins multiplying both embedding vectors to obtain a learnable projection matrix, which serves as a guide of how to mix the information coming from each dimension of the vectors. This matrix is then flattened and combined with path context (former stored projection matrices). A final linear layer is introduced to reshape the vector to its final form. At the end of this block, we have a combined vector with the information aggregated from two sources, keeping only the relevant features for the final goal.

Last, the new shared vector with the information combined from all sources feeds the five linear layers of the decoder. As an output, \textit{FAWN} gives the information of what is in the room (i.e., a robot and/or a person), and its localization.

\bigskip

This architecture allows \textit{FAWN} to leverage information from multiple sources, independently of its radio technology. Specifically in our testbed, it is able to effectively merge information from 5G and Wi-Fi networks, reaching greater accuracy levels than its individual parts, as we are going to see in the following section. 

\section{Results}
\label{sec:results}

In this section we evaluate the \textit{FAWN}'s performance. First set of experiments are devoted to analyze the classification ability. Last, we study the error on inducing the position of the target. For this purpose, we reuse the same scenario and data described in Sec.~\ref{sec:testbed}.
We choose \textit{CNN-ABLSTM}~\cite{he_robust_2023} as the best competitor from all tried articles. Although more recent work has come out (like \textit{WiSigPro}~\cite{hussain_wisigpro_2024}), we have seen that their performance is poorer within our scenario (see columns \textit{Position accuracy} and \textit{Scenario accuracy} of Table~\ref{tab:sota}). We also compare \textit{FAWN} against its two reduced solutions -- \textit{Wi-Fi only} and \textit{5G only} -- which are constructed using the same layers as \textit{FAWN}, but feeding the opposite encoder with null data (5G and Wi-Fi, respectively).

\textit{FAWN} training was conducted over 100 epochs, allowing sufficient iterations for convergence while preventing overfitting. To fulfill the training, we use Adam optimizer with learning rate of $10^{-3}$ and Cross-Entropy loss function. 
The entire process took approximately 30 mins, including both training and validation phases, over a NVIDIA DGX H200 GPU from the SLICES-Madrid Infrastructure. We can observe in Fig.~\ref{fig:training_results} the average performance over the epochs of \textit{FAWN} and its competitor \textit{CNN-ABLSTM}~\cite{he_robust_2023} while training. The average performance metric weights among classification accuracy and position estimation, showing the superiority of \textit{FAWN}. 

%
%

\begin{figure}%
    \centering%
    \begin{tikzpicture}

\definecolor{darkestblue}{RGB}{36,90,143}
\definecolor{darkblue}{RGB}{70,140,192}
\definecolor{ourblue}{RGB}{126,169,209}
\definecolor{lightblue}{RGB}{170,189,221}
\definecolor{lighterblue}{RGB}{209,209,231}
\definecolor{lightestblue}{RGB}{241,238,246}

  \begin{axis}[
        width=\columnwidth, 
        height=0.5\columnwidth,
        ylabel={Average performance}, 
        ymin=0.75, ymax=0.94, xmin=0, xmax=149,
        tick style={draw=none},
        xlabel={Epochs},
        grid=both,
        legend style={fill=none, draw=none, at={(.5,1)}, anchor=south, /tikz/every even column/.style={column sep=1em}},
        legend columns=2,
        ]

    \addplot[name path=upperFAWN, draw=none, forget plot] table[x=epoch, y expr=\thisrow{accuracy_avg} + \thisrow{accuracy_std}, col sep=comma] {csvs/training_performance_multi_encoder_isac.csv};
    
    \addplot[name path=lowerFAWN, draw=none, forget plot] table[x=epoch, y expr=\thisrow{accuracy_avg} - \thisrow{accuracy_std}, col sep=comma] {csvs/training_performance_multi_encoder_isac.csv};

    \addplot[fill=lightblue, draw=none, forget plot] fill between [of=upperFAWN and lowerFAWN];

    \addplot[name path=upperABLSTM, draw=none, forget plot] table[x expr=\thisrow{epoch}*5, y expr=\thisrow{accuracy_avg} + \thisrow{accuracy_std}, col sep=comma] {csvs/training_performance_cnn_ablstm.csv};
    
    \addplot[name path=lowerABLSTM, draw=none, forget plot] table[x expr=\thisrow{epoch}*5, y expr=\thisrow{accuracy_avg} - \thisrow{accuracy_std}, col sep=comma] {csvs/training_performance_cnn_ablstm.csv};

    \addplot[fill=lightblue, draw=none, forget plot] fill between [of=upperABLSTM and lowerABLSTM];

    \addplot [darkestblue, mark=x, mark size=6pt, mark repeat=25, mark phase=0, mark options={line width=2pt}, line width=1pt] table [x=epoch, y=accuracy_avg, col sep=comma] {csvs/training_performance_multi_encoder_isac.csv};
    \addlegendentry{\textit{FAWN}}
    
    \addplot [darkblue, mark=*, mark size=2pt, mark repeat=5, mark phase=0, mark options={line width=2pt}, line width=1pt] table [x expr=\thisrow{epoch}*5, y=accuracy_avg, col sep=comma] {csvs/training_performance_cnn_ablstm.csv};
    \addlegendentry{\textit{CNN-ABLSTM}~\cite{he_robust_2023}}


    
\end{axis}%
%

%
%
%
\end{tikzpicture}%
    \caption{Average performance of the training in function of the epochs. Light blue areas are the standard deviation.}%
    \label{fig:training_results}%
\end{figure}%

To evaluate the classification performance, we choose three well known metrics: the F1-Score, harmonic mean between the precision and the recall; the accuracy in predicting the presence of a human, a robot, or both; and the confidence of such decision. 
By looking at Fig.~\ref{fig:ECDF}-(top), we can see that \textit{FAWN} has the best performance against its competitor \textit{CNN-ABLSTM}~\cite{he_robust_2023} in F1-Score (0.951 vs. 0.836) in accuracy (0.960 vs. 0.867) and confidence (0.991 vs. 0.840). 
We can also appreciate the advantage of leveraging information from both bands, as reduced solutions \textit{Wi-Fi only} and \textit{5G only} perform worse than \textit{FAWN}, the last being able to effectively fuse the information.

Fig.~\ref{fig:ECDF}-(top) also indicates the number of parameters of each solution, and the mean error (measured in meters) when inferring the position. This latter result is expanded in Fig.~\ref{fig:ECDF}-(bottom), where the \gls{ecdf} of the position error in meters is plotted. We can appreciate that approximately 84\% of the samples of \textit{FAWN} have an error less than $0.6\,$m and almost 90\% have less than 1.2$\,$m, surpassing the rest of solutions. Notice that the step-like pattern shown by the curves correspond to the final grid layer outcome from the model and not due to a lack of executions. The counterparty comes when we observe the number of parameters. Although they are bigger in the case of FAWN, it is justified in terms of performance.

\begin{figure}[t]
\centering
{\footnotesize
\setlength{\tabcolsep}{3.3pt}
\begin{tabular}{lccccc}
\toprule
\hspace{1em}Model & F1-score & Accuracy & Confidence & Param. & \begin{tabular}{c} Mean \\ Error (m)  \end{tabular} \\ 
\midrule
\begin{tikzpicture}
\definecolor{darkestblue}{RGB}{36,90,143}
    \draw[darkestblue, mark=x, mark size=5pt, mark repeat=2, mark phase=2, line width=1.8pt] plot coordinates {(0,0) (.25,0) (.5,0)};
\end{tikzpicture}\textit{FAWN} & \textbf{0.951} & \textbf{0.960} & \textbf{0.991} & 1.22M & \textbf{0.26} \\
\begin{tikzpicture}
\definecolor{darkblue}{RGB}{70,140,192}
    \draw[darkblue, mark=*, mark size=1.66pt, mark repeat=2, mark phase=2, line width=1.8pt] plot coordinates {(0,0) (.25,0) (.5,0)};
\end{tikzpicture}\cite{he_robust_2023} & 0.836 & 0.867 & 0.840 & 469K & 0.42 \\
\begin{tikzpicture}
\definecolor{ourblue}{RGB}{126,169,209}
    \draw[ourblue, mark=triangle, mark size=1.66pt, mark repeat=2, mark phase=2, line width=1.8pt] plot coordinates {(0,0) (.25,0) (.5,0)};
\end{tikzpicture}\textit{Wi-Fi only} & 0.868 & 0.855 & 0.802 & \textbf{149K} & 0.77 \\
\begin{tikzpicture}
\definecolor{lighterblue}{RGB}{209,209,231}
    \draw[lighterblue, mark=square*, mark size=1.66pt, mark repeat=2, mark phase=2, line width=1.8pt] plot coordinates {(0,0) (.25,0) (.5,0)};
\end{tikzpicture}\textit{5G only} & 0.662 & 0.673 & 0.385 & 686K & 0.87 \\
\bottomrule
\end{tabular}}

\begin{tikzpicture}

\definecolor{darkestblue}{RGB}{36,90,143}
\definecolor{darkblue}{RGB}{70,140,192}
\definecolor{ourblue}{RGB}{126,169,209}
\definecolor{lightblue}{RGB}{170,189,221}
\definecolor{lighterblue}{RGB}{209,209,231}
\definecolor{lightestblue}{RGB}{241,238,246}

  \begin{axis}[
        width=\columnwidth,
        height=0.5\columnwidth,
        ylabel={ECDF},
        ymin=0.55, ymax=1,
        xmin=-0.2, xmax=7.5,
        tick style={draw=none},
        xlabel={Error (m)},
        ytick={.6,.7,.8,.9,1},
        grid=both,
        xtick={0,1,2,3,4,5,6,7,8,9,10},
        xticklabels = {0,0.6,1.2,1.8,2.4,3,3.6,4.2,4.8,5.4,6},
        ]
        
    \addplot [darkestblue, mark=x, mark size=6pt, mark repeat=40, mark phase=0, line width=2pt] table [x=error, y=cdf, col sep=comma] {csvs/fawn_ourmodel_cdf.csv};
    
    \addplot [darkblue, mark=*, mark size=2pt, mark repeat=40, mark phase=0, line width=2pt] table [x=error, y=cdf, col sep=comma] {csvs/wifi_sota_cdf.csv};

    \addplot [ourblue, mark=triangle, mark size=2pt, mark repeat=40, mark phase=0, line width=2pt] table [x=error, y=cdf, col sep=comma] {csvs/wifi_ourmodel_cdf.csv};

    \addplot [lighterblue, mark=square*, mark size=2pt, mark repeat=40, mark phase=20, line width=2pt] table [x=error, y=cdf, col sep=comma] {csvs/3gpp_ourmodel_cdf.csv};
    
  \end{axis}
\end{tikzpicture}
\caption{Results of the experimental evaluation of the models. (top) Average metrics. (bottom) \glspl{ecdf} of the position error in meters.} 
\label{fig:ECDF}
\end{figure}

To further analyze the behavior of the models, we conduct an experimental evaluation of the position error per location within the testing area.  
Fig.~\ref{fig:combined_heatmaps} depicts a top view representation of the room in which the measurements were taken, per each of the evaluated models. The figure shows the precision error regarding the positions, the darker the color, the greater the average inference error. White boxes corresponds to furniture, such us tables or shelves. 
As expected, areas where \textit{5G only} fails but \textit{Wi-Fi only} hits, \textit{FAWN} also hits (coordinate 4,1.2), and vice-versa. However, this representation allows us to prove that \textit{FAWN} is able to learn even on the areas where both \textit{5G only} and \textit{Wi-Fi only} have problems (coordinate 4,5.4), demonstrating that the fusion of the information goes beyond the simple addition of the result of both.


These results demonstrate the utility of combining different sources of information, being able to surpass literature solutions focused on extracting as much as possible from only one source. Notice here that \textit{FAWN} encoders may be optimized to further increase the knowledge they obtain from the environment, however, the objective of this section is to demonstrate the potential of the fusion of information. 



Even though these experiments were conducted over a single scenario, it is enough to
demonstrate \textit{FAWN}'s strong performance due to sensor fusion across the evaluated tasks, consistently outperforming baseline approaches in terms of accuracy and robustness. 
The model’s ability to effectively integrate multi-sensor data and focus on the most relevant features highlights its practical utility in real-world applications.

\begin{figure*}
\centering
\input{img/combined_heatmap}
\caption{Mean localization error map comparing \textit{5G}, \textit{Wi-Fi}, \textit{CNN-ABLSTM}, and \textit{FAWN}; top view representation of the testing area. The darker the color, the greater the average inference error. White boxes corresponds to furniture.}
\label{fig:combined_heatmaps}
\end{figure*}

\section{Conclusion and Future Line directions}
\label{sec:futureworks}

This work has illustrated the potential of using multi-technology wireless sensing, specifically combining Wi-Fi and \gls{5g} signals, to infer detailed environmental information in realistic indoor scenarios. 
The presented multi-encoder architecture of \textit{FAWN} has effectively fused heterogeneous data sources in a unified framework for a common localization task.


Several direction lines involve deploying task-specific decoders tailored to different use cases, such as activity recognition, safety monitoring, or semantic mapping. This modularity could further enhance the versatility of the system across diverse application domains. Additionally, we aim to explore the integration of a generative decoder, inspired by current large language models, to enable richer and more flexible outputs. Finally, expanding the framework to include temporal modeling techniques would strengthen the system’s capability to operate reliably in dynamic and evolving real-world environments.




\section*{Acknowledgments}

This work has been partially funded by the European Commission Horizon Europe SNS JU MultiX (GA 101192521) Project, and by the Regional Government of Comunidad de Madrid under grant agreement No. TEC-2024COM-360 (Disco6G); and it is part of the Project 6GINSPIRE PID2022-137329OB-C42, funded by MCIN/ AEI/10.13039/501100011033/. We would like to thanks SLICES-Madrid for their services (\url{https://www.slices-madrid.eu/funding-acknowledgements/}).




\bibliographystyle{IEEEtran}
\bibliography{references}

\newpage

 



\begin{IEEEbiographynophoto}{Carlos Barroso-Fernández}
got his M.Sc. in 2022 and is a Ph.D. student at Universidad Carlos III de Madrid.
\end{IEEEbiographynophoto}\vspace{-22pt}

\begin{IEEEbiographynophoto}{Alejandro Calvillo-Fernandez}
got his M.Sc. in 2024 and is a Ph.D. student at Universidad Carlos III de Madrid.
\end{IEEEbiographynophoto}\vspace{-22pt}
\begin{IEEEbiographynophoto}{Antonio de la Oliva}
 got his M.Sc. in 2004
and his Ph.D. in 2008, is an associate professor at Universidad Carlos III de Madrid.
\end{IEEEbiographynophoto}\vspace{-22pt}
\begin{IEEEbiographynophoto}{Carlos J. Bernardos}
 got his M.Sc. in 2003
and his Ph.D. in 2006 and he is a professor at Universidad Carlos III de Madrid.
\end{IEEEbiographynophoto}\vspace{-22pt}

\vfill

\end{document}